\documentclass[letterpaper, 10 pt, conference]{ieeeconf}  

\IEEEoverridecommandlockouts                              
\usepackage{xcolor}
\usepackage{comment}
\usepackage{graphics} 
\usepackage{epsfig} 
\usepackage{amsmath} 
\usepackage{amssymb}  
\usepackage{mathtools}
\usepackage{marginnote}
\usepackage{hyperref}
\usepackage{cite}
\usepackage{mathrsfs}
\usepackage{algorithm,algorithmic}
\usepackage{xcolor}
\usepackage[inkscapelatex=false]{svg}
\usepackage{multirow} 
\usepackage{array} 
\usepackage{etoolbox}
\usepackage[thicklines]{cancel}
\usepackage[inkscapelatex=false]{svg}
\usepackage{flushend}
\usepackage{booktabs}

\makeatletter
\patchcmd{\@makecaption}
  {\scshape}
  {}
  {}
  {}
\patchcmd{\@makecaption}
  {\\}
  {.\ }
  {}
  {}
\makeatother

\title{\LARGE \bf
DASH Robot: Minimalistic Design and Optimal Aerial-Terrestrial Locomotion via Contact-Implicit Control}
\author{
Ryan Gomes Paiva$^{1,*}$, Conrad Ho$^{1,*}$, Jiarong Kang$^{1,*}$, Kunzhao Ren$^{1}$, Xiangru Xu$^{1}$, and Xiaobin Xiong$^{1,2}$
\thanks{$^{*}$Equal contribution.}%
\thanks{$^{1}$University of Wisconsin--Madison, WI, USA.}%
\thanks{$^{2}$X. Xiong is now with the Shanghai Innovation Institute (SII), Shanghai, China,
and was with the University of Wisconsin--Madison.}%
\thanks{Corresponding to X. Xiong (\tt\small xiaobin.xiong@sii.edu.cn)}}
\begin{document}

\newcommand{\Kang}[1]{{\color{blue} \textbf{TODO: Kang #1}}}

\newcommand{\TODOR}[1]{{\color{cyan} \textbf{TODO Ryan: #1}}}

\newcommand{\TODOC}[1]{{\color{red} \textbf{TODO Conrad: #1}}}
\newcommand{\TODO}[1]{{\color{yellow} \textbf{TODO: #1}}}

\newcommand{\block}[1]{\noindent{\textbf{#1}:}}
\newcommand{\emphhh}[1]{{\color{yellow} \textbf{#1}}}

\maketitle
\pagestyle{empty}

\begin{abstract}
 We present a novel and minimalistic design of an aerial-terrestrial robot DASH: Ducted Aerial Spring Hopper. The goal is to enable both aerial and ground locomotion capabilities on a unified mobile robot that is mechanically-minimalistic, locomotion-versatile, and energy-efficient. We propose an organic integration of ducted fan co-axial body with a springy leg at the bottom for realization. The ducted fan module provides thrust-vectoring as the main actuation for agile flying; when it is combined with the light-weight spring leg, the robot realizes highly efficient ground hopping with energy circulation. Moreover, to realize optimal locomotion with two modes, we employ a contact-implicit model predictive controller to automatically choose locomotion modes and actuation. We successfully validated the design and control of DASH through a range of tasks, including periodic hopping, aerial flight, and mode-free locomotion with autonomous mode transitions during obstacle traversal.
\end{abstract}

\section{Introduction}




Mobile robots are increasingly expected to operate across diverse environments that include both structured and unstructured terrains. Legged robots \cite{wensing2023optimization, dai2022bipedal} are efficient for traversing irregular ground but struggle with discontinuities and obstacles that exceed their reach. Aerial robots can bypass such obstacles but incur high energetic cost and limited endurance \cite{float, coaxial-duct}. These complementary strengths motivate hybrid aerial–terrestrial robots that combine the efficiency of legs with the agility of flight \cite{hybrid-mit, pogox, hybridrobot, flybot}. Such systems promise to extend mobility beyond what is possible with either modality alone, enabling applications in search-and-rescue \cite{siegwart2015legged}, inspection \cite{alexis2016aerial}, and exploration \cite{agha2021nebula}.

\begin{figure}
    \centering
    \includegraphics[width=1.0\linewidth]{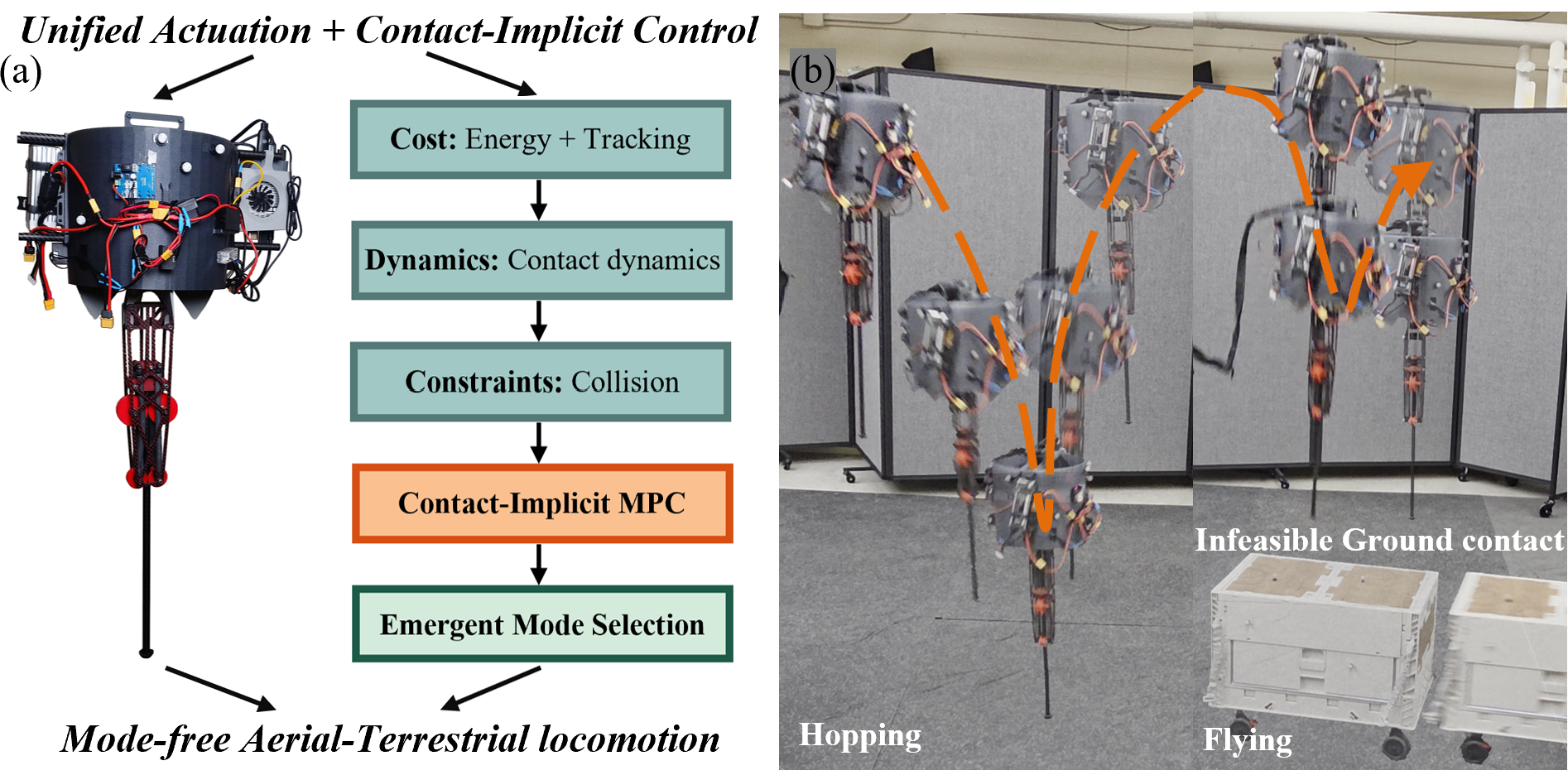}
    \vspace{-15pt}
    \caption{(a) The hybrid robot DASH, equipped with unified actuation, addresses the challenge of mode-free aerial-terrestrial locomotion through Contact-Implicit Control. (b) DASH autonomously transitions between hopping and flight in response to environmental constraints, balancing feasibility and optimality.}
    \vspace{-15 pt}
    \label{fig:first}
\end{figure}

We introduce a ducted fan-legged robot, DASH — Ducted Aerial Spring Hopper, that is capable of both hopping and flying as illustrated in Fig. \ref{fig:first}. The design integrates a spring-loaded leg for terrestrial hopping \cite{pogox, flybot} with ducted fans \cite{coaxial-duct} for vertical thrust and assisted aerial maneuvers. This combination enables the robot to traverse flat terrain, negotiate cluttered environments, and cross large gaps by seamlessly transitioning between ground and air modes. Compared to other state-of-the-art aerial-terrestrial robots that have open-propeller systems, such as PogoX \cite{pogox} and PogoDrone \cite{zhu2022pogodrone}, DASH with the ducted fan \cite{kim2023design} offer high thrust-density, compactness, enhanced safety, and potential capabilities to perform efficient physical interactions with the environment. Additionally, the duct acts as a guard against debris or foreign objects entering the blades, enhancing durability. 


Despite the promise of hybrid platforms, locomotion control remains a central challenge. Robot dynamics inherently couples continuous evolution with discrete contact events such as stance, liftoff, and landing. Conventional approaches have mainly focused on the control realization of individual modes, biased toward terrestrial locomotion \cite{pogox}. The exploration of multi-modal locomotion typically relies on predefined mode schedules \cite{zhu2022pogodrone, zeng2024reference, SkateDuct}, which restricts optimality since all possible contact sequences must be anticipated in advance. Another common approach uses layered designs \cite{hu2023trajectory, gherold2025self, 10016673, suh2020energy} that combine sampling-based planning and low-level control. However, lower-dimensional models are used in the planning, which restricts the optimal solution from being found or dynamic constraints from being respected in dynamic maneuvers. 

To address these challenges, we adopt the Contact-Implicit Control (CIC) \cite{kim2025contact, kong2023hybrid, le2024fast, jiang2024contact, esteban2025reduced} in the legged robotics and manipulation community to solve the planning and control problem on multi-domain locomotion. The CIC eliminates the need to manually specify contact sequences or preselect between flying and hopping modes for aerial-terrestrial robots. Therefore, CIC allows the robot to automatically determine when and where contacts occur. Locomotion behaviors thus emerge directly from solving the control problem rather than from handcrafted logic. For DASH, this capability is critical: it enables seamless transitions between hopping and flight without explicit mode switching, under a unified cost design shaped around energy circulation.

We demonstrate that together, these contributions highlight how contact-implicit control, coupled with ducted-fan robot design, expands the mobility and versatility of hybrid aerial–terrestrial robots in environments requiring seamless hybrid operation.

\section{Related Work}

\block{Aerial-Terrestrial Robots} The early literature on aerial-terrestrial robots has focused on enabling wheeled locomotion \cite{ramirez2025multimodal, 10380664, suh2020energy} on flying vehicles. Recent studies have also explored the potential of enabling legged locomotion with flying. PogoX \cite{pogox} and Hopcopter \cite{flybot} demonstrated hopping–flying systems that exploit low thrust-to-weight ratio and elastic energy storage to improve terrestrial efficiency and durability. These ideas were extended to insect-scale platforms \cite{hybrid-mit}. Other embodiments include quadrotor-assisted bipeds \cite{zeng2025kou} and flying–crawling quadrotors \cite{hu2023trajectory}, showing the diversity of system-level innovations at a preliminary stage in hybrid aerial-terrestrial locomotion.

\block{Ducted Fan Robot} Ducted fans on aerial vehicles \cite{mccormick1999aerodynamics} and ducted fan vehicles \cite{coaxial-duct, kim2023design, 6244694} have been well studied and used on aerial vehicles as they can offer advantages over open rotors, including higher thrust efficiency, reduced tip losses, and safer operation in cluttered environments. Yet, they have rarely been applied to robotics with a few exceptions, such as enabling flying of DRAGON \cite{zhao2018design} and bipedal robots \cite{9340714}. The control of ducted-fan vehicles typically requires several control surfaces to direct the thrust direction for attitude control. Similar ideas have been widely explored on aerial vehicles \cite{float} to enable agile flying maneuvers. That said, ducted fans have not yet been applied to hybrid mobile robots despite their potential benefits over open propellers.



\block{Control of Hybrid Robots} On the control side, flying and ground controllers are typically designed and implemented separately. Flying behaviors are typically implemented via linear controllers \cite{zhu2022pogodrone, flybot}, while ground locomotion, such as hopping, is typically implemented via step-level stepping controllers \cite{pogox} on top of height controllers. Besides, traditional model predictive control (MPC) has been applied to wheeled aerial robot \cite{10016673}, and state-of-the-art data-driven predictive control \cite{zeng2024reference} has demonstrated high-accuracy locomotion of flying and hopping. At the planning level, self-supervised cost-of-transport estimation \cite{gherold2025self} offers a scalable way to optimize multimodal trajectories, complementing earlier work on energy-efficient hybrid motion planning using sampling-based methods \cite{suh2020energy}. Yet, to the best of our knowledge, optimal control of hybrid locomotion across multiple domains with different dynamics remains underexplored.

\block{Contact-Implicit Control} Planning and control around contact has been an essential topic in legged locomotion and robotic manipulation. Common approaches often utilize soft-contact modes \cite{neunert2018whole}, alternative smoothing of vector fields \cite{westenbroek2021smooth}, or complementarity conditions of contact \cite{pang2023global} to make contact implicit in the optimization of planning and control. Complementarity-based formulations of rigid-body contact dynamics \cite{leins2025mujoco, le2024contact} have recently been widely adopted in Contact-Implicit MPC (CI-MPC) frameworks \cite{kim2025contact, kong2023hybrid, le2024fast} to address legged locomotion and manipulation problems. However, the potential of CI-MPC for planning across diverse locomotion modes in hybrid mobile robots remains largely unexplored, which constitutes a central focus of this work.

\section{Robot Design}

DASH is a compact hybrid robot that integrates thrust-assisted aerial locomotion with spring-loaded terrestrial hopping. Its architecture combines a ducted fan for lift and thrust vectoring with a pogo-stick-inspired leg for ground interaction. To support these functions, the system is built around lightweight structural components, high-power actuation, and onboard computation for real-time control.




    

\subsection{Mechanical Design}

DASH combines a ducted fan propulsion unit with a pogo-stick spring leg to enable both aerial and terrestrial locomotion. The central body houses the fan assembly, enclosed by a 3D-printed duct using PLA-Aero for safety, aerodynamic efficiency, and structural support. The fan is composed of two motorized coaxial propellers that are mounted in the duct via lightweight carbon-fiber tubes and 3D-printed PLA-CF materials. Attached to the bottom of the duct are 3D-printed vanes that provide thrust vectoring for stabilization and maneuverability. The ducted body is designed with a 10-inch radius to provide sufficient moment authority through thrust vectoring. 

The leg is designed as a pogo-stick spring mechanism, enabling DASH to store and release elastic energy for efficient hopping maneuvers. The elastic tubing spring also reduces the thrust required for fast takeoffs when flying from hopping. A pair of spaced pulleys constrains the leg’s vertical motion, providing smooth translation while minimizing slack compared to a single linear bearing. This novel integration of the ducted fan with the spring leg allows DASH to combine energy-efficient terrestrial motion with thrust-assisted aerial capabilities.

\begin{figure}
    \centering
    \includegraphics[width=0.85\linewidth]{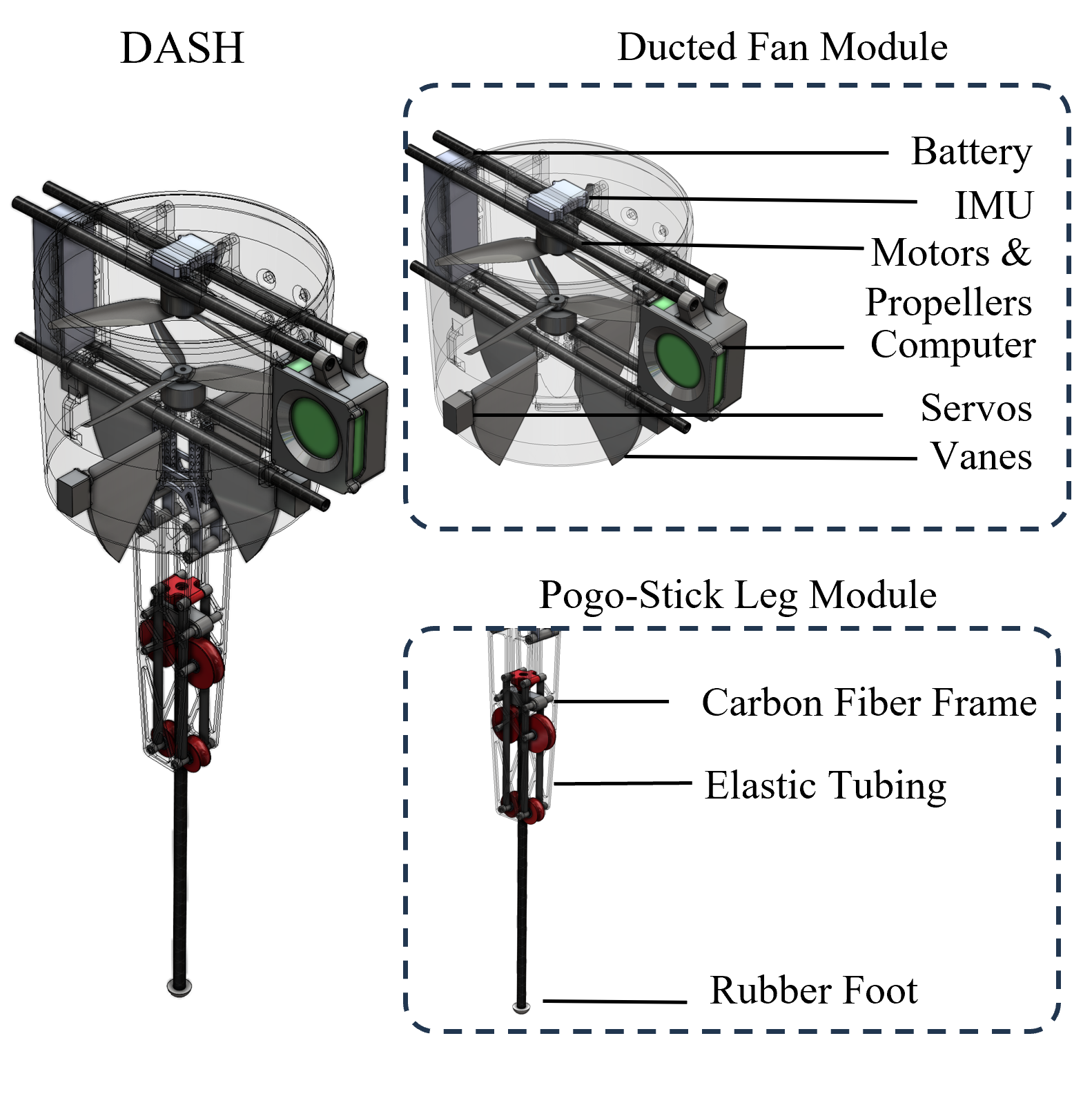}
        \vspace{-10pt}
    \caption{Illustration of the mechanical components of DASH. }
    \label{fig:robotdesign}
\end{figure}

The total structural weight of the 3D-printed duct body and vanes is 719 g, balancing robustness with manufacturability. Together with propulsion and onboard electronics, the total system mass is approximately 3.27 kg, with a height of 4 feet. A schematic of the robot design is shown in Fig. \ref{fig:robotdesign}, highlighting the integration of the pogo-stick spring leg with the ducted fan assembly. 


\begin{table}[h!]
\centering
\begin{tabular}{llr} 
\specialrule{1.5pt}{0pt}{0pt} 
\textbf{Component} & \textbf{Model} & \textbf{Weight (g)} \\
\specialrule{0.8pt}{0pt}{0pt} 
Battery & HRB 6S 3300mAh 60C & 467 \\
Motor & T-Motor Cine77 3610 977KV & 400 \\
Propeller &  Gemfan 1050 Reinforced Nylon & 34 \\
ESC & V-Good RC 32-Bit 100A  & 148 \\
Flight Controller & Holybro Pixhawk 6C Mini  & 39 \\
IMU & x-io x-IMU3 & 50 \\
Servo & Dynamixel XC330-T181-T & 92 \\
Computer & UP-Squared-i12 & 111 \\
Duct Body & 3D Print & 609 \\
Vanes & 3D Print & 110 \\
Leg & Carbon Fiber & 542 \\
Wires & Harness & 670 \\
\specialrule{0.8pt}{0pt}{0pt} 
\textbf{Total} & & \textbf{3272} \\
\specialrule{1.5pt}{0pt}{0pt}
\end{tabular}
\caption{Component list with models and weights.}
\label{tab:component_specs}
    \vspace{-10pt}
\end{table}

\subsection{Electronics and Actuation}

The electronics of DASH are designed to provide reliable high-power actuation alongside real-time computation for hybrid locomotion. The ducted fan is driven by two coaxial brushless DC motors, while thrust vectoring is realized through four servo actuators mounted on the duct. Together, these actuators generate lift and directional control for both flight and thrust-assisted terrestrial maneuvers.

Onboard computation is distributed across two systems. A PX4 flight controller manages low-level motor regulation, ensuring stable thrust generation and reliable actuator communication. A single-board computer (SBC) equipped with an Intel i7 processor executes higher-level planning and control in real time. The robot is equipped with a top-mounted IMU for orientation estimation, while translational states are obtained from a motion capture system. The system can be extended to fully onboard visual–inertial state estimation using a camera, as shown in related work \cite{MHEFast}.

\begin{figure}
    \centering
     \includegraphics[width=1\linewidth]{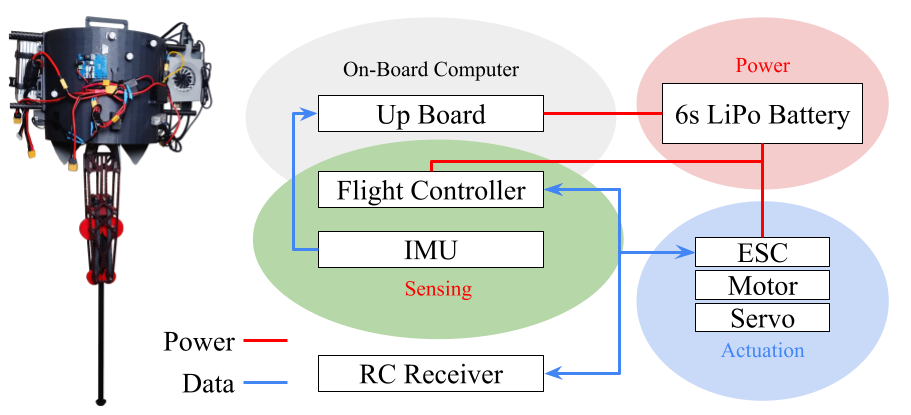}
    \caption{Illustration of the electronics architecture of DASH. }
        \vspace{-15pt}
    \label{fig:robotelectronics}
\end{figure}

 All the electronics are powered by one 6S LiPo battery that supplies both propulsion and computation, with distribution designed to isolate high-current actuation from sensitive electronics. Figure. \ref{fig:robotelectronics} summarizes the electronic architecture, while Table \ref{tab:component_specs} provides a breakdown of the weights.

The PX4 flight controller provides low-level motor communication and PWM/servo actuation, while attitude regulation is performed by the geometric controller. The SBC supplies the computational resources for higher-level planning and control. For hybrid locomotion, DASH generates motor PWM commands and desired orientation trajectories through the CI-MPC. Both thrust and orientation are planned within the Contact-Implicit MPC framework, with thrust allocation playing a primary role in contact transitions and mode switching. The PX4 executes the PWM commands, while a quaternion-based geometric attitude controller \cite{AttitudeCtrl} tracks the planned orientation. This architecture allows us to evaluate the proposed contact-implicit control strategy without relying on off-the-shelf flight stabilization.

\section{Robot Dynamics}
We now present the dynamic modeling of DASH. Owing to the coexistence of parameterized actuation mechanisms and hybrid locomotion behavior, we first describe the \textit{actuation dynamics} and their key parameters, followed by the formulation of the hybrid \textit{locomotion dynamics}.

\subsection{Actuation Model}
\begin{figure}[t]
    \centering
    \includegraphics[width=0.96\linewidth]{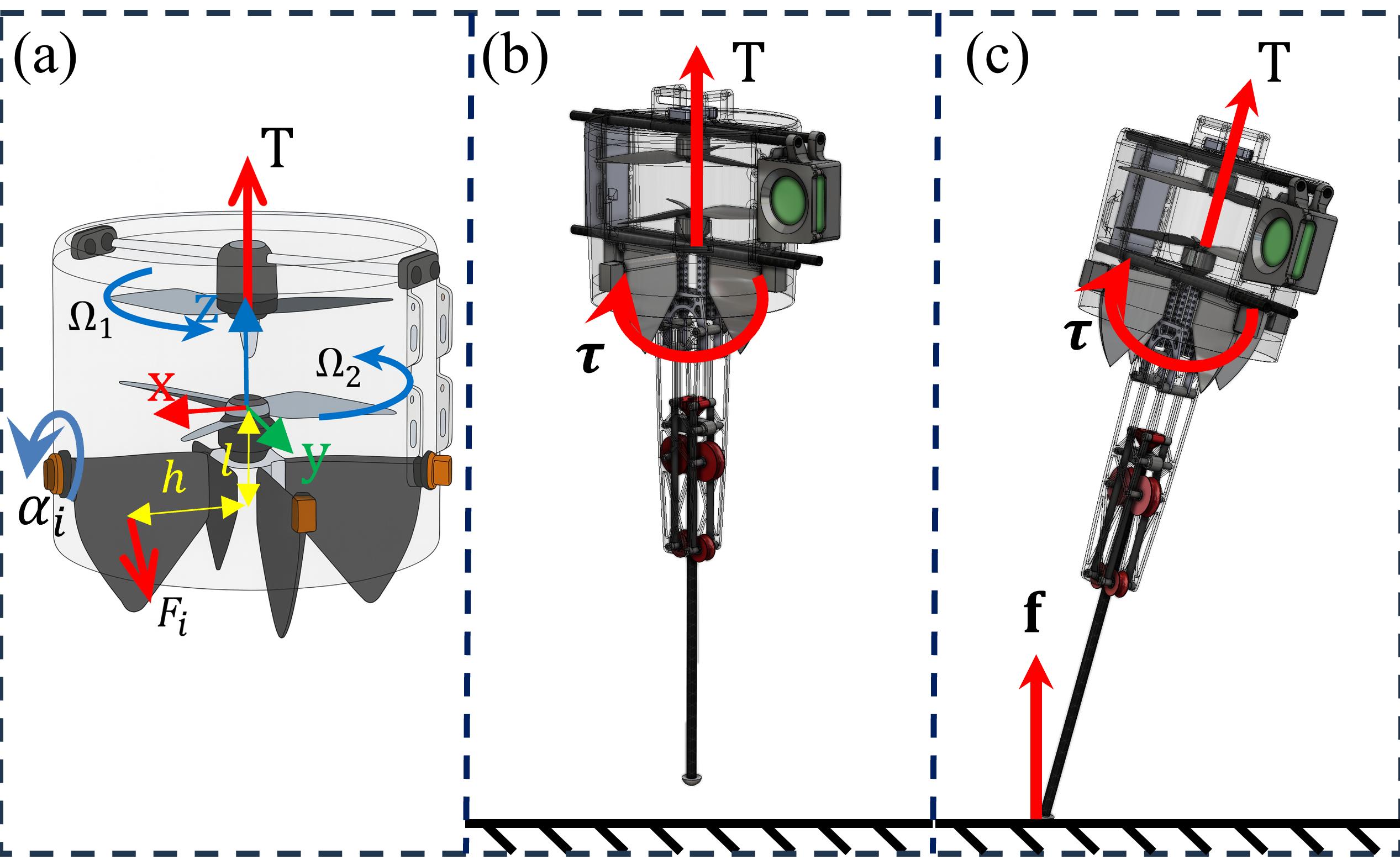}
    \vspace{-5pt}
    \caption{(a) The actuation of DASH, (b) the aerial dynamics being actuated by the total thrust and moment, and (c) the ground dynamics with additional ground reaction force.}
        \vspace{-15pt}
    \label{fig:dynamicsModel}
\end{figure}

DASH is actuated by the total thrust force  $\mathbf{T} \in \mathbb{R}^3$ and moment $\boldsymbol{\tau} \in \mathbb{R}^3$ generated from the propellers and vanes, as indicated in Fig. \ref{fig:dynamicsModel}. We first examine how the thrusts and moments are generated based on aerodynamics.

\block{Thrust Force Model}  We adopt common assumptions on flying robots that the total thrust from two propellers is: 
\begin{equation}
     \mathbf{T}_p =  [ 0, 0, c_1 \Omega^2_1 + c_2\Omega^2_2]^\intercal,\label{eq:thrustforce}
\end{equation}
where $\Omega$ denotes the rotational velocity of each propeller, and $c_1$ and $c_2$ are constant coefficients determined by the propellers and physical dimensions of the ducted body.

\block{Moment Model of Co-Axial Motors} The two propellers also create a turning moment in the body frame z-axis: 
\begin{equation}
    \boldsymbol{\tau}_{p} = [0, 0, c_3(\Omega_1^2 - \Omega^2_2)]^\intercal,\label{eq:coaxial_moment}
\end{equation}
where $c_3$ is a parameter related to the propellers.  

\block{Vane Force Model} The vanes direct the airflow beneath the ducted fan. The change in airflow results in aerodynamic push force $F_i$ on the vane $i$, where $i = 1,2,3, \text{or } 4$ is the index of the vanes. The push force $F_i$ is modeled as: 
\begin{equation}
    F_i = \frac{1}{2} \rho A C V^2_o \alpha_i \label{eq:vaneforce},
\end{equation}
where $\rho$ is the air density, $A$ is the vane surface area, $C$ is a constant coefficient, $V_o$ denotes the air speed at the duct outlet, and $\alpha$ represents the servo angle. Here, small angle approximation of $\alpha$ is used. The air speed $V_o^2$ is assumed proportional to the total thrust force magnitude $T_p$. Thus, \eqref{eq:vaneforce} can be rewritten as: $F_i = \frac{1}{2} \rho A c_4 T_p \alpha_i,$
with $c_4$ being another constant coefficient. 

\block{Total Force} The total thrust is obtained by summing the thrust of the propeller and the contributions of the four vane forces: $\mathbf{T} = \mathbf{T}_p + \sum^4_{i = 1} \mathbf{F}_i$.


\block{Moment} $\mathbf{T}_p$ and gravitational force go through the center of mass (COM). Therefore, aside from the propeller reaction moment $\boldsymbol{\tau}_{p}$, the remaining moments are generated by the vane forces:
$\boldsymbol{\tau} = \mathbf{r} \times \mathbf{F}$. Assuming small angles that the vane force location is approximately fixed, the moment in the body frame is a linear relation to the vane angles: 
\begin{equation}
    \boldsymbol{\tau}_v = \frac{1}{2} \rho A c_4 T_p  \begin{bmatrix}
        -l & 0 & l & 0 \\
        0 &  -l & 0 & l \\
      h & h & h & h 
    \end{bmatrix}  \begin{bmatrix}
        \alpha_1 \\ \alpha_2\\ \alpha_3\\ \alpha_4
    \end{bmatrix},
\end{equation}
where $l$ and $h$ represent the distances from the aerodynamic center of the vane to the body-frame x-axis and z-axis, respectively.
The total moment thus is $\boldsymbol{\tau} =\boldsymbol{\tau}_p + \boldsymbol{\tau}_v.$

The parameters $c_1, c_2, c_3,  c_4$ in the above equations are identified through hardware experiments, which are detailed in Section VI. 

\subsection{Hybrid Locomotion Dynamics}
During locomotion, the robot is modeled as a single rigid body with a springy leg. Thus, we model the robot with 7 degrees of freedom.

\block{Aerial Dynamics}
The leg is designed to be naturally stable due to the spring. Additionally, the leg and foot are of small weight. As a result, we do not model the dynamics of the leg within the robot in aerial motion. Assuming the robot is a single rigid body that is actuated by the total thrust $\mathbf{T}$ and moment $\boldsymbol{\tau}$ from the propellers and vanes, its dynamics are:
\begin{align}
    \mathbf{I} \dot{\boldsymbol{\omega}} + \boldsymbol{\omega} \times ( \mathbf{I} \boldsymbol{\omega}) = \boldsymbol{\tau},  \label{eq:eulerEq} \\
    m \dot{\boldsymbol{\mathit{v}}} = \mathbf{R} \cdot \mathbf{T} + m\mathbf{g},
\end{align}
where $\mathbf{I} \in \mathbf{R}^{3\times3}$ is the lumped body inertia of the robot, $m$ is its total mass, $\mathbf{R}\in SO(3)$ represents the body orientation, $\boldsymbol{\mathit{v}}$ is the linear velocity, and $\boldsymbol{\omega}$ is the angular velocity.

\block{Terrestrial Dynamics}
When the robot is on the ground, the leg with its spring provides forces to its upper body as internal actuation. The robot is modeled as a regular legged robot, with a floating-base coordinate and a prismatic springy joint. Let $\mathbf{q}\in SE(3) \times \mathbb{R}^1$ denote its configuration, and let $\mathbf{v} \in \mathbb{R}^7$ denote its velocity. Its dynamics with a holonomic contact constraint are: 
\begin{align}
    \mathbf{M}(\mathbf{q}) \dot{\mathbf{v}} + \mathbf{h}(\mathbf{q}, \mathbf{v}) =  \mathbf{u}_r +  \mathbf{J}(\mathbf{q})^\intercal  \mathbf{f}, \label{eq:lag_dyn}\\
     \mathbf{J}(\mathbf{q}) \dot{\mathbf{v}} + \dot{ \mathbf{J}}(\mathbf{q},\mathbf{v}) {\mathbf{v}} = 0,
\end{align}
where $\mathbf{M}(\mathbf{q})$ is the mass matrix, and $\mathbf{h}(\mathbf{q}, \mathbf{v})$ denotes the Coriolis and gravitational vector, $\mathbf{u}_r $ denotes the robot actuation, $\mathbf{J}(\mathbf{q})$ represents the Jacobian matrix at the foot contact, and $\mathbf{f}$ denotes the ground reaction force. For DASH, $ \mathbf{u}_r =  [\mathbf{T}^\intercal, \boldsymbol{\tau}^\intercal,  0]^\intercal$. 

At events such as landing (contact established) or lift-off (contact released), the system undergoes discrete transitions. These are captured by an impact map that instantaneously updates the velocity:  
\begin{align}
    \mathbf{M}(\mathbf{q})(\mathbf{v}^{+} - \mathbf{v}^{-}) &= \mathbf{J}(\mathbf{q})^\intercal \boldsymbol{\lambda}, \\
    \mathbf{J}(\mathbf{q}) \mathbf{v}^{+} &= \mathbf{0},
\end{align}
where $\mathbf{v}^-$ and $\mathbf{v}^+$ are the pre- and post-impact velocities, and $\boldsymbol{\lambda}$ denotes the impulse at contact.  
Similar hybrid dynamics models have been commonly used in legged locomotion \cite{grizzle2014models}.

\section{Contact-Implicit Control}

\begin{figure}
    \centering
    \includegraphics[width=1.0\linewidth]{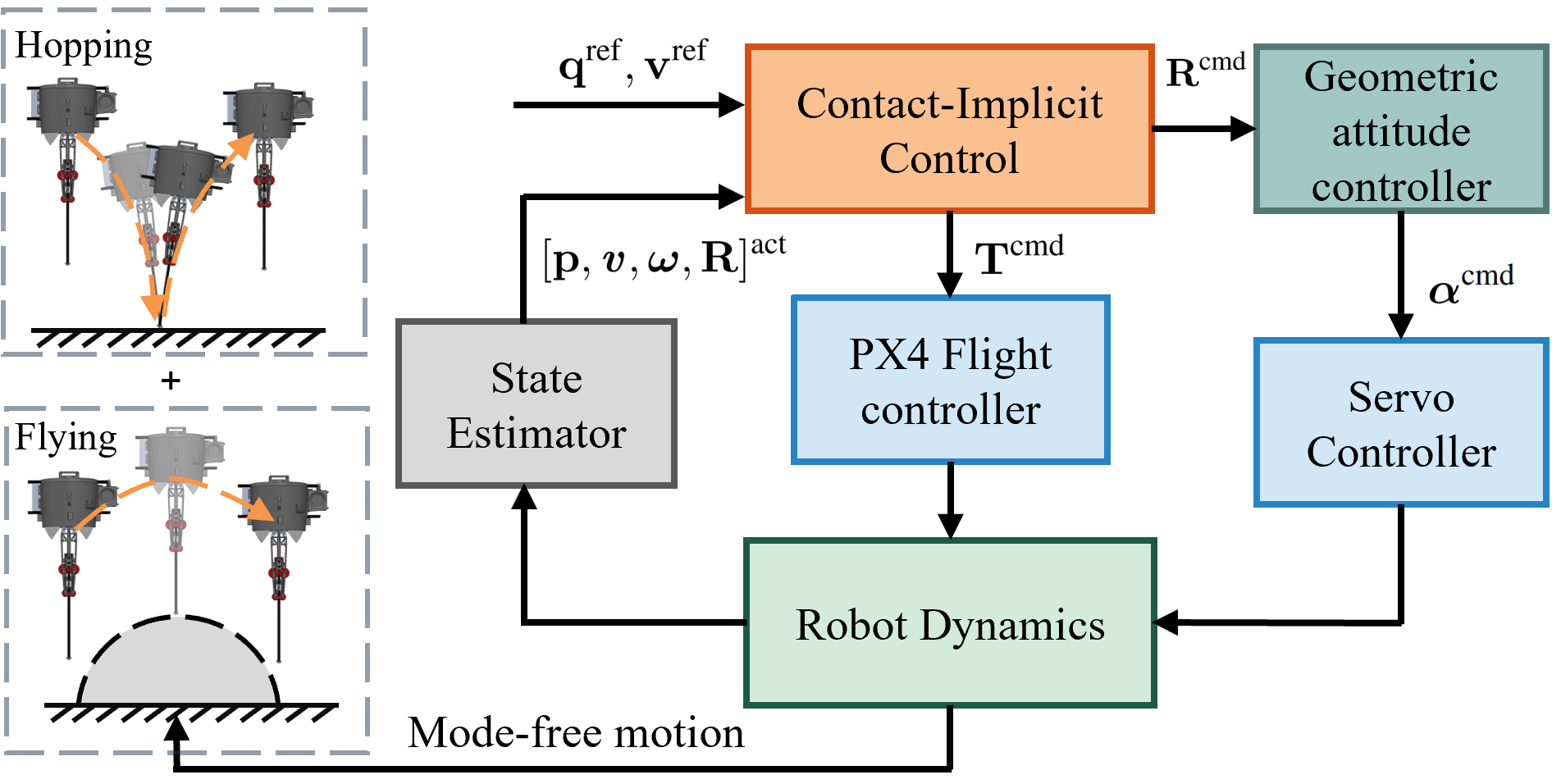}
    \vspace{-15pt}
    \caption{Control architecture of DASH enabling mode-free motion planning.}
    \vspace{-15pt}
    \label{fig:diagram}
\end{figure}

Now we present our control formulation to realize multi-modal locomotion on this robot. We choose the thrust force and rotational moment as the input to the robot: $\mathbf{u} = [\mathbf{T}^\intercal, \boldsymbol{\tau}^\intercal,0]^\intercal$. The system state is $\mathbf{x} = [\mathbf{q}^\intercal, \mathbf{v}^\intercal]^\intercal.$ The core of contact-implicit control lies in the modeling of contact dynamics. While the hybrid dynamics formulation described previously is well suited for theoretical analysis, it complicates controller design. In conventional approaches, locomotion behaviors are typically constructed by manually scheduling contact sequences and switching between predefined modes. For example, terrestrial locomotion \cite{pogox} is often designed independently, with translational velocity regulated through step-to-step dynamics and height controlled via a nonlinear controller. The transitions between aerial and terrestrial modes are hardcoded \cite{SkateDuct,zeng2024reference}. Such designs restrict hybrid robots from achieving truly mode-free motion planning. To address this limitation, we adopt complementarity-based contact dynamics within a model predictive control framework, unifying planning and control in a recursive manner. This formulation, as shown in Fig. \ref{fig:diagram}, allows the robot to compute optimal control actions directly from the dynamics, without relying on manually defined locomotion modes or predefined switching logic.


\subsection{Complementarity-Conditioned Dynamics}
Without prescribing the contact sequence in advance, we model the contact dynamics using the following complementarity conditions. 

\noindent{\textbf{Time-stepping Dynamics:}}  
For rigid body systems modeled by Lagrangian dynamics, the continuous time dynamics \eqref{eq:lag_dyn} can be approximated by a discrete-time system using a semi-implicit Euler integration scheme:
\begin{align}
    \mathbf{q}^+ - \mathbf{q} &= \Delta t\, \mathbf{v}^+, \\
    \mathbf{M}(\mathbf{q})({\mathbf{v}}^{+} - {\mathbf{v}}) &= \Delta t \big(\mathbf{u}_r - \mathbf{h}(\mathbf{q}, \mathbf{v}) + \mathbf{J}(\mathbf{q})^\intercal \mathbf{f} \big), \label{eq:discrete_dyn}
\end{align}
where $\mathbf{q}^+$ and $\mathbf{v}^+$ denote the state at the next time step, and $\Delta t$ is the discrete time interval. Instead of imposing regularized dynamics for predefined contact modes, the contact impulse $\boldsymbol{\lambda} = \Delta t \mathbf{f}$ is obtained from a complementarity formulation of the contact constraints, enabling mode-consistent contact resolution.

\noindent{\textbf{Non-penetration:}}  
For contact at the next time step, both the signed distance between two bodies $\phi(\mathbf{q}^{+})$ and the normal contact impulse $\lambda^n = \Delta t f^{n}$ during the interval must be non-negative. Moreover, normal contact impulse can only be nonzero when two bodies are in contact: $\phi(\mathbf{q}^{+}) \geq 0 \perp \lambda^{n} \geq 0.$
A linear approximation of this condition is used for computation \cite{CEKF}:
\begin{equation}
    \phi(\mathbf{q}) + \nabla \phi(\mathbf{q}) (\mathbf{q}^{+} - \mathbf{q}) \geq 0 \perp \lambda^n \geq 0. \label{eq:linear_complementarity}
\end{equation}

\noindent{\textbf{Maximum Dissipation:}}  
The maximum dissipation principle states that tangential friction impulses act to maximize the dissipation of kinetic energy. In our single foot-contact scenario, the Coulomb friction model is approximated by a polyhedral cone, yielding the following formulation:
\begin{align} 
    \min_{\boldsymbol{\beta}} \quad &\boldsymbol{\lambda}^\intercal \mathbf{J}(\mathbf{q}) {\mathbf{v}}^+ \label{eq:max_disp} \\ 
    \text{s.t.} \quad &\boldsymbol{\lambda} = \mathbf{n} \lambda^n + \mathbf{D} \boldsymbol{\beta}, \ \boldsymbol{\beta} \geq \mathbf{0}, \label{eq:lamda_decompose}\\ 
    &\mu \lambda^n - \mathbf{e}^\intercal \boldsymbol{\beta} \geq 0. \label{eq:polyhedral_cone}
\end{align}
where $\mathbf{n}$ is the unit normal vector at the contact point, and $\mathbf{D}$ positively spans the tangential contact plane with unit vectors. The vector $\boldsymbol{\beta}$ represents the friction force coefficients along the directions defined by $\mathbf{D}$, $\mu$ is the coefficient of friction, and $\mathbf{e}$ is a vector of ones. The Karush–Kuhn–Tucker (KKT) conditions of the associated minimization problem are expressed as the following complementarity constraints:
\begin{align}
    \mathbf{D}^\intercal \mathbf{J} \mathbf{v}^{+} + \mathbf{e} \eta \geq \mathbf{0} & \perp \boldsymbol{\beta} \geq \mathbf{0},\\
    \mu \lambda^n - \mathbf{e}^T \boldsymbol{\beta} \geq 0  &\perp \eta \geq 0,
\end{align}
where \( \eta \) denotes the Lagrange multiplier. The KKT conditions, together with the discrete-time robot dynamics~\eqref{eq:discrete_dyn} and the non-penetration constraint~\eqref{eq:linear_complementarity}, give rise to a linear complementarity problem (LCP), whose solution determines the next state under contact dynamics:
\begin{equation}
    \mathbf{x}^{+} = \textbf{LCP}(\mathbf{x}, \mathbf{u}).
    \label{eq:contact_dynamics}
\end{equation}
where the notation indicates that the state transition is defined through the solution of the linear complementarity problem. The LCP-based contact dynamics admit analytical gradients, which can be exploited to improve computational efficiency in optimization-based algorithms \cite{kim2025contact}.

\subsection{Model Predictive Control}
We apply model predictive control (MPC) to the dynamics of DASH formulated as a LCP \eqref{eq:contact_dynamics}. The MPC regulates the torso motion while planning contact interactions by balancing optimality and feasibility under the imposed costs and constraints. At each step, a nonlinear program is solved to determine an optimal sequence over a horizon of $N$ steps. Only the first control input is executed, and the process is repeated at the next time step.

\block{Cost Function}
The cost $\mathcal{J}_{\text{MPC}}$ includes trajectory tracking, control regularization, and mechanical energy tracking:
\begin{equation}
\begin{aligned}
\mathcal{J}_{\text{MPC}} 
=& \textstyle \sum_{k=0}^{N}
\big(
\| \mathbf{x}_k - \mathbf{x}_{\mathrm{ref},k} \|_{\mathbf{Q}_{\mathbf{x}}}^{2}
\\ & + \| E(\mathbf{x}_k) - E_{\mathrm{ref}} \|_{Q_E}^{2}
\big) + \textstyle \sum_{k=0}^{N-1}
\| \mathbf{u}_k \|_{\mathbf{R}}^{2},
\end{aligned}
\end{equation}
where $\mathbf{Q}_{\mathbf{x}}$, $Q_E$, and $\mathbf{R}$ denote the weights associated with each cost term, $E(\mathbf{x}_k)$ denotes the mechanical energy of the system. The energy term enforces coherent energy circulation across contact and flight phases, supporting consistent hybrid behavior. The relative weighting between trajectory tracking and control effort determines how energy is regulated: stronger control penalization promotes passive energy exchange through terrain interaction and compliance, whereas weaker penalization allows greater active energy injection. As a result, distinct locomotion behaviors emerge from the optimization without explicit mode scheduling.

\block{Constraints}
The system dynamics at each time step are enforced as constraints, together with an initial state condition and control input bounds. Obstacle avoidance is enforced through a minimum-distance constraint $ \| \mathbf{p}_{\text{frame}} - \mathbf{p}_{\text{obs}} \| \geq \alpha $, which guarantees a separation of at least \( \alpha \) between the robot frame position \( \mathbf{p}_{\text{frame}} \) and the obstacle position \( \mathbf{p}_{\text{obs}} \).

\block{MPC formulation} The MPC is concisely represented as:
\begin{align*}
&\textstyle \min_{\{\mathbf{x},\mathbf{u}\}_{k=0}^{N}} \quad  \mathcal{J}_{\text{MPC}} &&\tag{MPC} \label{eq:MHE}\\
\text{s.t.} \quad 
& \mathbf{x}_0 = \tilde{\mathbf{x}}_0, \tag{Initial State}\\
&\mathbf{x}_{k+1} = \textbf{LCP}(\mathbf{x}_k, \mathbf{u}_k),  \tag{Dynamics}\\
&\mathbf{0}  \leq \textbf{Constr}(\mathbf{x}_k, \mathbf{u}_k), \tag{Constraints} \end{align*}
where the bottom constraint represents all the state and input constraints. 

\subsection{Solution Method via Differential Dynamic Programming}
At each control step, we solve a finite-horizon MPC subproblem using Differential Dynamic Programming (DDP) \cite{crocoddyl}. DDP enforces dynamics implicitly by rolling out the state trajectory from the current system state and optimizes only the control sequence. The update is performed through a local quadratic approximation of the Bellman value and Q-functions, combined with a backward–forward iteration. This produces a time-varying affine feedback law over the horizon that tracks the reference trajectory while respecting the LCP-based dynamics model. Since the LCP dynamics are only piecewise differentiable, the analytic gradients are relaxed to improve numerical performance \cite{kim2025contact}.

\block{Relaxed Analytic Gradient} 
Taking into account the contact mode associated with the current iterated trajectory, the contact mode is classified into three types based on the contact complementarity constraints: no contact, clamping contact $(\cdot)_c$, and sliding contact $(\cdot)_s$.
Using the specified contact modes, the original LCP problem is transformed into a set of linear velocity-level constraints:
\begin{equation}
    \begin{bmatrix} \boldsymbol{\mathit{v}}_c^n \\ \boldsymbol{\mathit{v}}_c^t \\ \boldsymbol{\mathit{v}}_s^n \end{bmatrix} 
    = \begin{bmatrix} \mathbf{0} \\ \mathbf{0} \\ \mathbf{0} \end{bmatrix} 
    = \begin{bmatrix} \mathbf{J}_c^n \\ \mathbf{J}_c^{t} \\ \mathbf{J}_s^n \end{bmatrix} \mathbf{v}^+,
\end{equation}
where \( \boldsymbol{\mathit{v}} \) denotes the contact-frame velocity, and \( (\cdot)^{n} \) and \( (\cdot)^{t} \) represent the normal and tangential components of the impulse, frame velocity, or Jacobian, respectively. The resulting contact force is then expressed as:
\begin{align}
        \begin{bmatrix}
        \boldsymbol{\lambda}^{n}_c \\
        \boldsymbol{\lambda}^{t}_c\\
        \boldsymbol{\lambda}^{n}_s
    \end{bmatrix} &= -\mathbf{A}_{cc}(\mathbf{q})^{-1} \mathbf{b}_{cc}(\mathbf{q},\dot{\mathbf{q}}), \\
\mathbf{A}_{cc} &=
\begin{bmatrix}
\mathbf{J}_c^n \\
\mathbf{J}_c^{t} \\
\mathbf{J}_s^n
\end{bmatrix}
\mathbf{M}^{-1}
\begin{bmatrix}
\mathbf{J}_c^n \\
\mathbf{J}_c^{t} \\
\mathbf{J}_s^n + \mathbf{J}_s^t \,\mu \,\operatorname{sign}(\boldsymbol{\lambda}_s^t)
\end{bmatrix}^{\intercal}, \\
\mathbf{b}_{cc} &=
\begin{bmatrix}
\mathbf{J}_c^n \\
\mathbf{J}_c^{t} \\
\mathbf{J}_s^n
\end{bmatrix}
\mathbf{M}^{-1}
\left[
\Delta t \,(\mathbf{u}_r - \mathbf{h}) + \mathbf{M} \mathbf{v}
\right].
\end{align}
Then, the gradient of the contact impulse is obtained as
\begin{equation}
    \partial 
    \begin{bmatrix}
        \boldsymbol{\lambda}_c^{n} \\
        \boldsymbol{\lambda}_c^{t} \\
        \boldsymbol{\lambda}_s^{n}
    \end{bmatrix}
    =
    \mathbf{A}_{cc}^{-1}
    \,\partial \mathbf{A}_{cc}\,
    \mathbf{A}_{cc}^{-1}
    \mathbf{b}_{cc}
    -
    \mathbf{A}_{cc}^{-1}
    \,\partial \mathbf{b}_{cc},
\end{equation}
where \( \partial \) denotes the total derivative with respect to an associated parameter vector.

The analytic gradient of the contact dynamics is computed within a local region that preserves the contact mode. To provide an informative gradient that enables contact breaking, the complementarity constraints of the contact foot in the normal direction: 
$\boldsymbol{\mathit{v}}^n \geq 0 \perp \boldsymbol{\lambda}^n \geq 0$,
is smoothed using a relaxation variable $\rho > 0$:
\begin{equation} 
\boldsymbol{\mathit{v}}^n \geq 0,\ \boldsymbol{\lambda}^n \geq 0, \ \boldsymbol{\mathit{v}}^n \boldsymbol{\lambda}^n = \rho. \end{equation}
This smoothing only takes effect in the differentiation, where velocity constraints in the normal direction are now expressed as:
\begin{equation}
    \frac{\rho}{\boldsymbol{\lambda}^n} = \mathbf{A}^n_{cc} \boldsymbol{\lambda}^n + \mathbf{b}^{n}_{cc},
\end{equation}
where the implicit relation between $\partial \boldsymbol{\lambda}^n$ and $\boldsymbol{\lambda}^n$ is given by:
\begin{equation}
    -\frac{\rho}{(\boldsymbol{\lambda}^n)^2} \partial\boldsymbol{\lambda}^n = \partial \mathbf{A}_{cc}^n \boldsymbol{\lambda}^n + \mathbf{A}_{cc}^n \partial \boldsymbol{\lambda}^n + \partial \mathbf{b}_{cc}. 
\end{equation}
Then the relaxed normal force gradient is given by:
\begin{equation}
    \partial \boldsymbol{\lambda}^n = -(\mathbf{A}_{cc}^n + \frac{\rho}{(\boldsymbol{\lambda}^n)^2})^{-1} (\partial \mathbf{A}_{cc}^n \boldsymbol{\lambda}^n + \partial \mathbf{b}_{cc}).
\end{equation}
This relaxed gradient facilitates smooth transitions between contact modes when incorporated into DDP~\cite{kim2025contact}.

\section{Results}
\label{sec:results}
We now present the results of controlling DASH. We first present the system identification of the robot, particularly the actuation model that is rarely seen on common legged robots. Then we show the control results of the robot in simulation and hardware performing different tasks. 

\block{System Identification} 
DASH generates complex aerodynamic forces through its propellers and vanes. Rather than identifying the coefficients $c_1$–$c_4$ from first principles, we estimate them experimentally. To identify $c_1$ and $c_2$, the robot is mounted on a single-DOF testing stand, similar to a seesaw, with a weight scale placed on the opposite side. By varying the BLDC motor speeds and measuring the resulting thrust forces, we obtain the corresponding coefficients. To identify $c_3$ and $c_4$, the ducted fan assembly is mounted on a 3-DOF rotational stand. By commanding specific servo angles to adjust the vanes and generating known thrust levels, we measure the resulting moments using a force gauge, from which the coefficients are estimated. In practice, unequal speeds of the coaxial counter-rotating propellers introduce nonlinear thrust--moment coupling. We therefore enforce \( \Omega_1 = \Omega_2 \), yielding \( \boldsymbol{\tau}_p = 0 \) and simplifying the model without compromising achievable behaviors.

\begin{figure}
    \centering
    \includegraphics[width=0.95\linewidth]{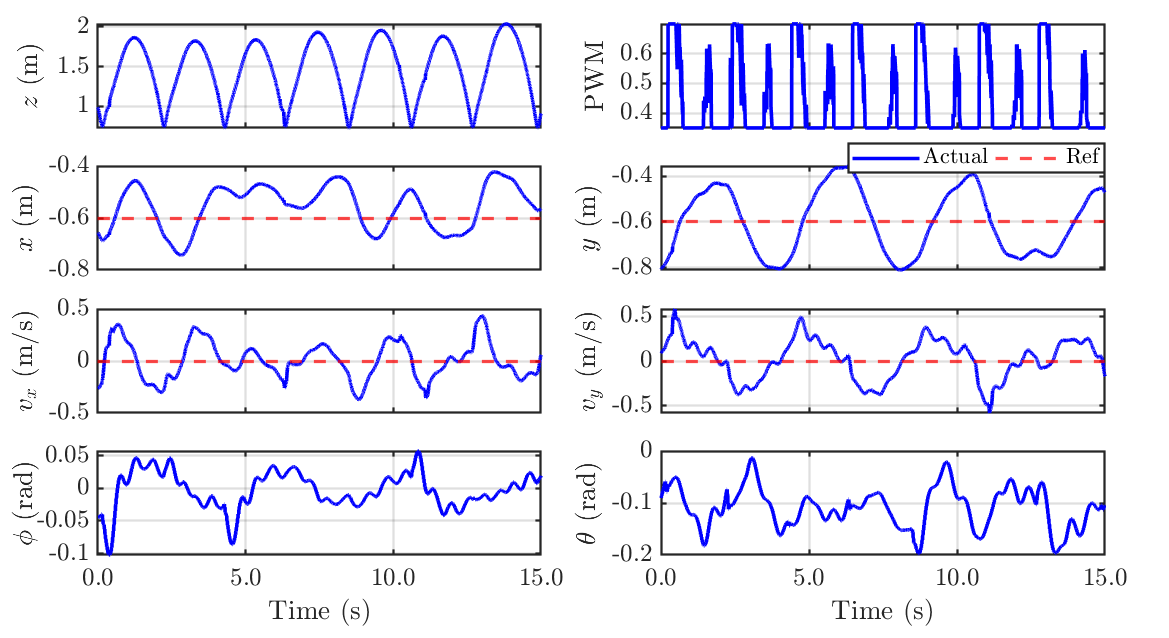}
    \vspace{-10pt}
    \caption{Experimental results of CI-MPC controlled in-place hopping.}
    \label{fig:hop_inplace}
    \vspace{-15pt}
\end{figure}

\block{Simulation and Software}
The robot is designed and assembled in SolidWorks, where each component is assigned its measured mass and geometric properties. This allows precise computation of the inertia and center-of-mass (COM) location. The robot is simulated using a time-stepping LCP-based contact dynamics model within a ROS2 environment. The controller and communication stack are fully implemented in C++, leveraging Crocoddyl \cite{crocoddyl} for efficient optimal control computation. The CI-MPC is configured with a 0.6\,s prediction horizon and a time discretization of 0.03\,s.
\begin{figure}
    \centering
    \includegraphics[width=0.80\linewidth]{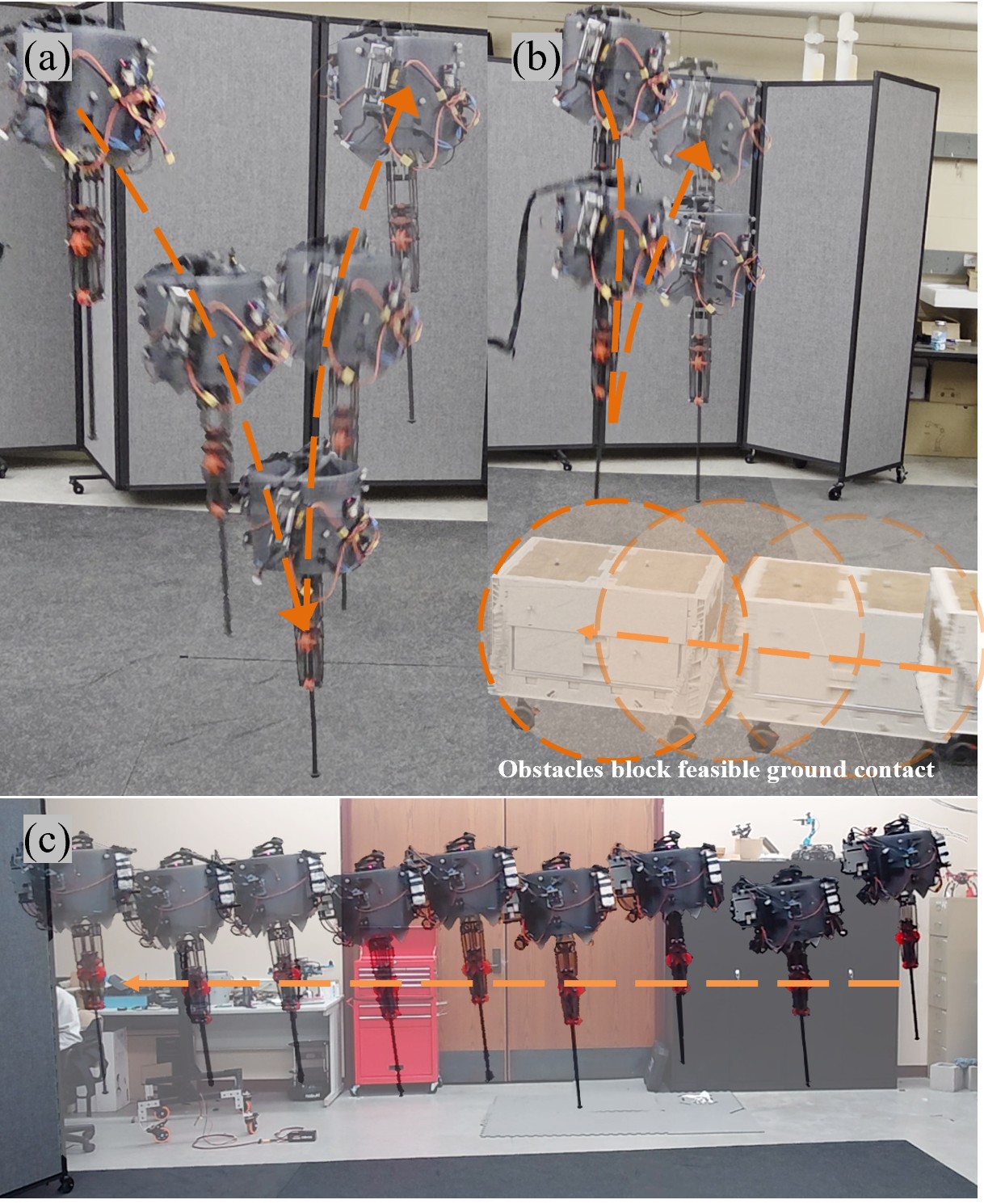}
    \vspace{-5pt}
    \caption{Experimental demonstration of DASH. (a, b) Transition from hopping to flight planned by the CI-MPC, arising from feasibility and optimality trade-offs. (c) Flying with velocity tracking of 0.5 m/s.}
    \vspace{-5pt}
    \label{fig:motion_exp}
\end{figure}
\begin{figure}
    \centering
    \includegraphics[width=1.0\linewidth]{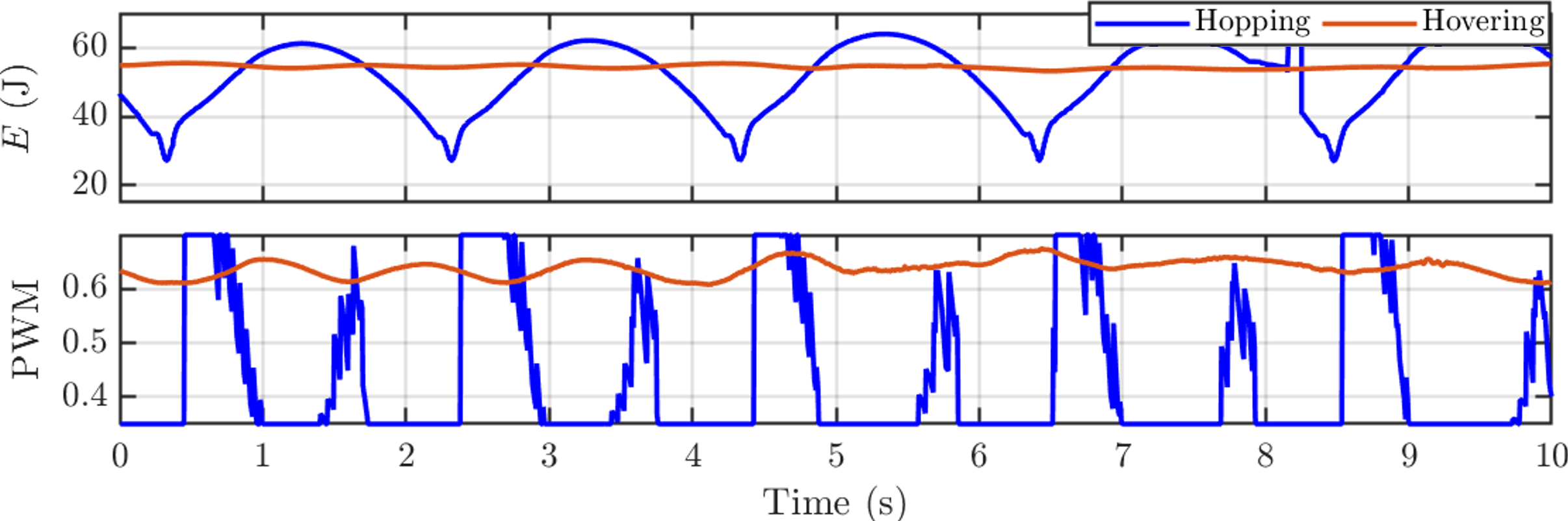}
    \vspace{-15pt}
    \caption{Comparison of mechanical energy circulation and control effort consumption in DASH during hopping and hovering.}
    \vspace{-15pt}
    \label{fig:energy}
\end{figure}

\subsection{Flying and Hopping}
In the flying experiments, the robot performs hovering with global position tracking and follows a reference linear velocity of up to 0.5\,m/s, constrained by the laboratory workspace as illustrated in Fig.~\ref{fig:motion_exp}.

In the hopping experiments, the robot executes continuous in-place hopping while tracking global position targets with zero reference translational velocity as shown in Fig.~\ref{fig:hop_inplace}. The resulting behavior maintains energy circulation while reducing control effort. Compared to flying, the optimized solution decreases the injected control effort, defined as $\int_{t_{\text{start}}}^{t_{\text{end}}} \left| \mathrm{T}(t) - \mathrm{T}_{\min} \right| \, dt$, to \(26.78\%\) of that in the flying case, as illustrated in Fig.~\ref{fig:energy}. This reduction reflects the efficient hybrid regime discovered by the CI-MPC. From Fig.~\ref{fig:energy}, it can be observed that the majority of the residual energy expenditure arises from the minimal thrust command required to keep the motors active, rather than from sustained thrust injection. Unlike hovering, which requires continuous active power to counteract gravity, hopping leverages intermittent energy storage and release through contact dynamics, thereby reducing active thrust usage and improving overall efficiency.
\subsection{Contact-Implicit Control}
\begin{figure}
    \centering
    \includegraphics[width=0.9\linewidth]{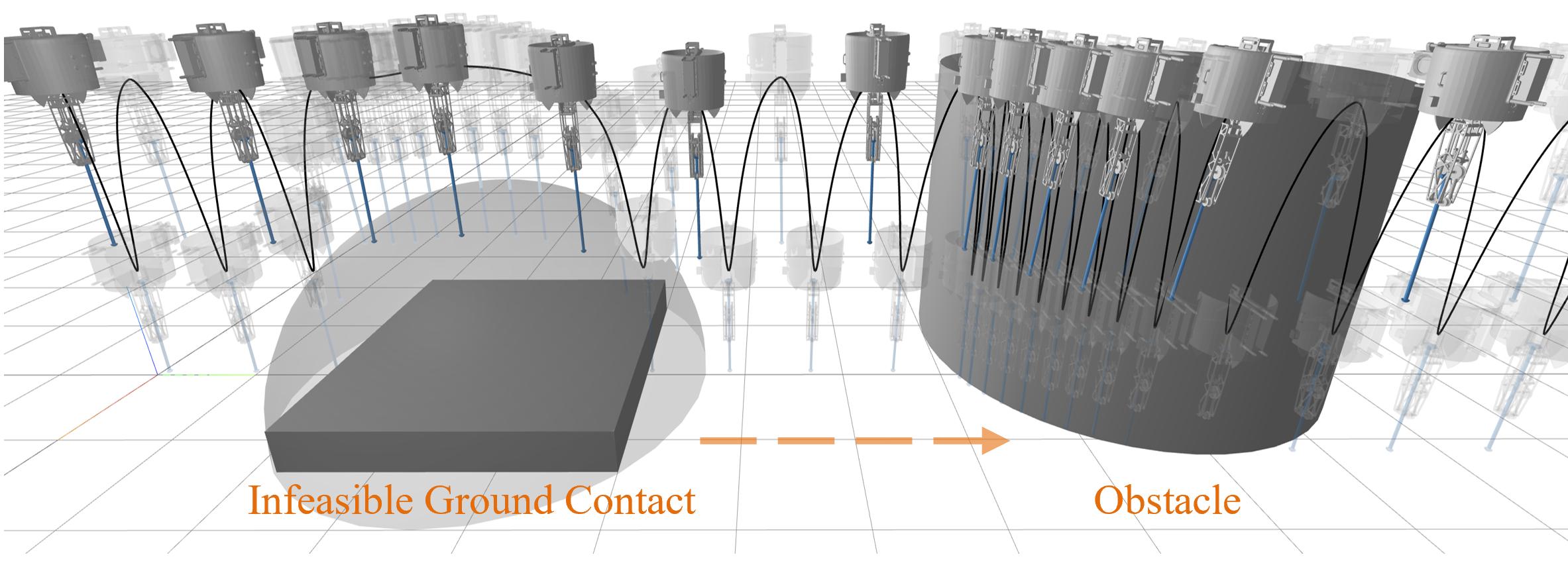}
    \vspace{-10pt}
    \caption{Simulation results of DASH tracking a 0.3 m/s velocity command, with hopping and flying emerging from feasibility–optimality trade-offs.}
    \vspace{-10pt}
    \label{fig:sim}
\end{figure}

\begin{figure}
    \centering
    \includegraphics[width=1.0\linewidth]{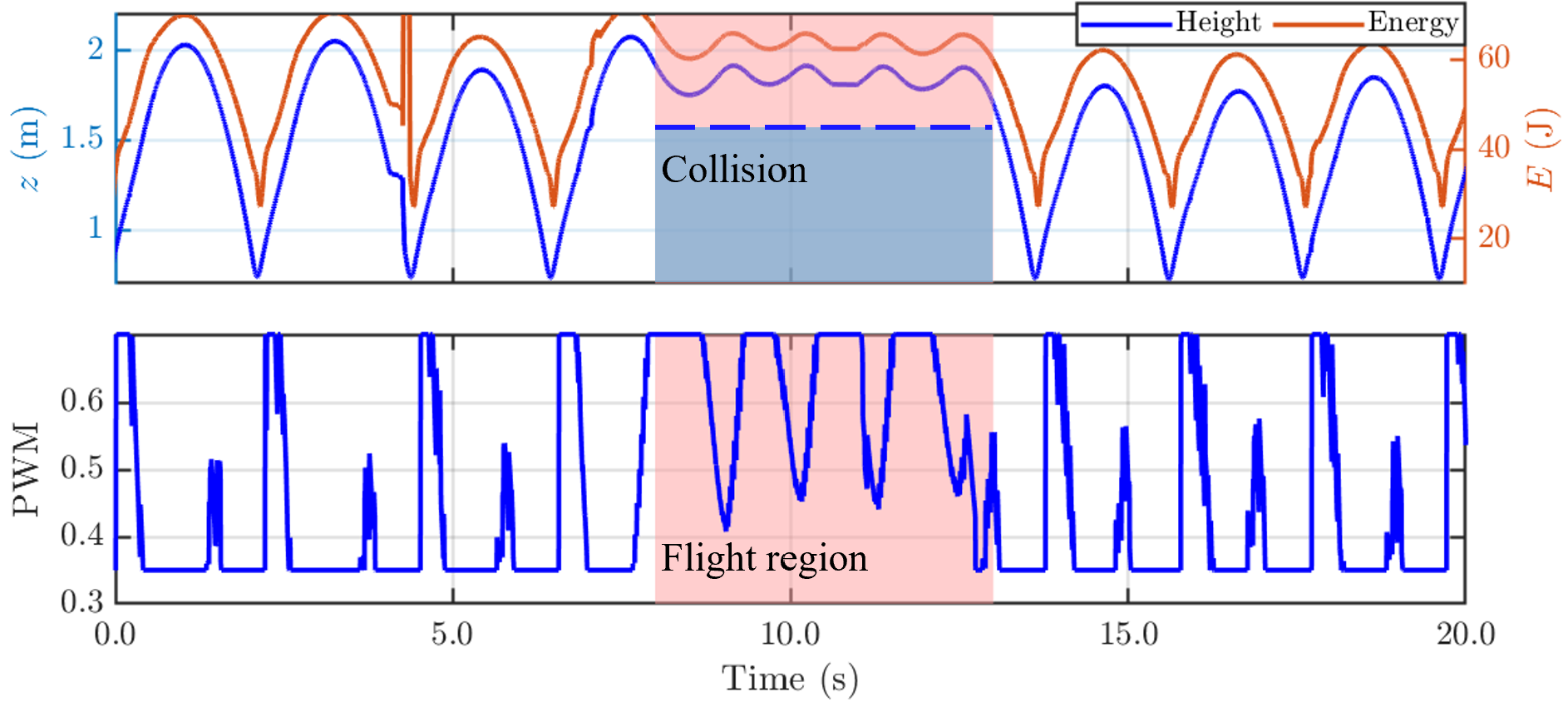}
    \vspace{-20pt}
    \caption{Hybrid locomotion generated by the CI-MPC on hardware, where obstacle-triggered collision constraints regularize the motion toward the flight region.}
    \vspace{-20pt}
    \label{fig:mode_switch}
\end{figure}

We present results of mode-free motion planning, in which the CI-MPC autonomously generates hopping behavior that minimizes control effort when ground contact is feasible. Simulation results are shown in Fig.~\ref{fig:sim}, where the robot traverses obstacles using mode-free locomotion. On hardware, when an obstacle appears beneath the robot, the activated collision avoidance constraints eliminate feasible contact configurations, causing the optimizer to transition the robot to flight, as illustrated in Fig.~\ref{fig:motion_exp}. This transition is not explicitly scheduled but instead emerges from feasibility considerations within the constrained optimization. From an energy perspective, when contact is feasible, the CI-MPC exploits passive energy exchange through terrain interaction and compliant leg dynamics, resulting in hopping with reduced active thrust injection. When contact becomes infeasible due to obstacle constraints, the optimizer shifts toward flight to maintain feasibility. Once the obstacle is removed and the collision constraints become inactive, ground contact becomes feasible again, and the CI-MPC naturally returns to the hopping regime, replanning the contact sequence to recover energy-efficient behavior, as shown in Fig.~\ref{fig:mode_switch}.

\section{Conclusion}
To conclude, we present a controller design that enables mode-free motion planning and control for the novel Ducted Aerial Spring Hopper (DASH). The proposed controller effectively integrates a contact-implicit control formulation with an energy-centered cost design. A key advantage of this framework is its ability to achieve automatic locomotion mode transitions by balancing feasibility and optimality within a unified optimization problem. With optimized performance in terms of energy circulation and control effort, the resulting hopping behavior reduces the injected control effort to \(26.78\%\) of that required for hovering in hardware experiments. This improvement arises from intermittent energy storage and release through contact dynamics, which reduces the need for sustained active thrust injection.

In future work, we aim to enable DASH to operate in outdoor and natural environments, where mode-free motion planning becomes increasingly critical. We are also interested in expanding the range of locomotion behaviors beyond the current capabilities, exploring additional hybrid modes such as swimming.

\bibliography{reference/ducted, reference/general,reference/hybridrobot, reference/contact_control}

\bibliographystyle{IEEEtran}

\end{document}